\documentclass[conference]{IEEEtran}
\IEEEoverridecommandlockouts
\usepackage{cite}
\usepackage{amsmath,amssymb,amsfonts}
\usepackage{algorithmic}
\usepackage{graphicx}
\usepackage{subcaption}
\usepackage{textcomp}
\usepackage{xcolor}

\def\x{{\mathbf x}}

\def\t{{\mathbf t}}

\def\BibTeX{{\rm B\kern-.05em{\sc i\kern-.025em b}\kern-.08em
    T\kern-.1667em\lower.7ex\hbox{E}\kern-.125emX}}
\begin{document}

\title{Interpretable Image Quality Assessment via CLIP with Multiple Antonym-Prompt Pairs
\thanks{This work was supported by research grant from the Telecommunications Advancement Foundation and JSPS KAKENHI Grant Number JP23K03871.}
}

\author{\IEEEauthorblockN{Takamichi Miyata}
\IEEEauthorblockA{
\textit{Chiba Institute of Technology}\\
Chiba, Japan \\
takamichi.miyata@it-chiba.ac.jp}
}

\maketitle

\begin{abstract}
  No reference image quality assessment (NR-IQA) is a task to estimate the perceptual quality of an image without its corresponding original image. It is even more difficult to perform this task in a zero-shot manner, i.e., without task-specific training.
  In this paper, we propose a new zero-shot and interpretable NR-IQA method that exploits the ability of a pre-trained vision-language model to estimate the correlation between an image and a textual prompt.
  The proposed method employs a prompt pairing strategy and multiple antonym-prompt pairs corresponding to carefully selected descriptive features corresponding to the perceptual image quality. Thus, the proposed method is able to identify not only the perceptual quality evaluation of the image, but also the cause on which the quality evaluation is based. 
  Experimental results show that the proposed method outperforms existing zero-shot NR-IQA methods in terms of accuracy and can evaluate the causes of perceptual quality degradation.
\end{abstract}


\section{Introduction}
No reference image quality assessment (NR-IQA) is a technology for estimating the perceptual quality of a degraded image without using its corresponding original image \cite{Moorthy2020-biqi,Saad2012-blinds2,Mittal2012-wz}.
Deep learning has become an invaluable tool in the field of NR-IQA in recent years. 
Although these methods effectively learn various types of distortions and their impact on the perceptual quality of images, they necessitate an extensive IQA-dataset, which consists of a large number of images and quality ratings, for training.

Contrastive language-image pretraining (CLIP) \cite{Radford2021-im} is a vision language model that has demonstrated its capability in zero-shot classification across various recognition tasks. Leveraging the zero-shot vision language capability of CLIP, Wang et al. have been introduced CLIP-IQA as a zero-shot NR-IQA method that can estimate the perceptual quality of a degraded image without needing to train on an IQA dataset \cite{Wang2022exploring}.

However, while CLIP-IQA can estimate the perceptual quality of an image, it does not provide insight into the basis for the quality judgment. In this paper, we propose a novel, zero-shot, and interpretable IQA method that utilizes several descriptive features potentially related to the perceived quality of an image.

Our approach offers an alternative to zero-shot IQA, focusing on multiple descriptive features. By developing several antonym-prompt pairs corresponding to carefully selected descriptive features and evaluating the similarity between these pairs and the image using CLIP, we can determine not only the perceived quality rating of the image, but also the factors influencing the rating.

\section{Proposed Method}

We have proposed a new zero-shot and interpretable IQA method using several descriptive features that are related to the perceived quality of an image.
The proposed method consists of two steps: (1) prompt pairing with multiple antonym-prompt pairs and (2) similarity measurement.

\subsection{Multiple antonym-prompt pairs}

Inspired by CLIP-IQA \cite{Wang2022exploring}, we use a prompt pairing strategy to estimate the perceptual quality of an image. 
The main limitation of CLIP-IQA is that it can only judge whether an image is good or bad, and cannot tell us from which perspective the quality of the image was judged to be good or bad.
To address this problem, we propose to use several antonym-prompt pairs corresponding to descriptive features related to the perceived quality of an image.
We use the following three descriptive features that correspond to perceptual image quality: sharpness, noise, and brightness.
Based on these descriptive features, we develop three antonym-prompt pairs as follows,
\begin{itemize}
    \item $[\t_{1,1}, \t_{1,2}]=[$``This is a good photo because it is sharp.'', ``This is a bad photo because it is blurred.''$]$
    \item $[\t_{2,1}, \t_{2,2}]=[$``This is a good photo because it is noiseless.'', ``This is a bad photo because it has noise.''$]$
    \item $[\t_{3,1}, \t_{3,2}]=[$``This is a good photo because it is light.'', ``This is a bad photo because it is dark.''$]$
\end{itemize}
where $\t_{i,j}$ is the $j$-th prompt ($j=1$ and $j=2$ correspond to positive and negative prompts, respectively) of the $i$-th antonym-prompt pair.

\subsection{Similarity measurement}
Let $\x$ be the input image, we first compute the similarity $s_{i,j}$ between the image $\x$ and the prompt $\t_{i,j}$  as follows,
\begin{equation}
    s_{i,j} = \frac{E_i(\x)^\top E_t(\t_{i,j})}{\parallel E_i(\x) \parallel_2^2+\parallel E_t(\t_{i,j}) \parallel_2^2 } \times 100,
\end{equation}
where $E_i$ and $E_t$ are the image and text encoder of CLIP \cite{Radford2021-im}, which maps the image/text into 1,024-dimensional vector in the common embedding space.

Then, we use softmax function to compute the descriptive scores $d_i$ from the similarity scores $s_{i,j}$ as follows,
\begin{equation}
    d_i = \frac{\exp(s_{i,1})}{\exp(s_{i,1}) + \exp(s_{i,2})} \times 100.
\end{equation}

The overall predicted perceptual quality score $q$ of the proposed method is the average of the descriptive scores from the all antonym-prompt pairs,
\begin{equation}
    q = \frac{1}{N}\sum_{i=1}^N d_i,
\end{equation}
where $N$ is the number of antonym-prompt pairs.

\section{Experimental Results}
We use CLIP \cite{Radford2021-im} with a ResNet-50 backbone as the vision language model.
Similar to CLIP-IQA \cite{Wang2022exploring}, we remove the positional embedding from the CLIP model to avoid the size limitation of the input image. 
We use KonIQ-10k \cite{Hosu2020-xg}, which is a dataset of 10,073 images with mean opinion score (MOS) annotations, for our experiment. To evaluate the performance of the proposed method, we use the Spearman's rank-order correlation coefficient (SROCC) and Pearson's linear correlation coefficient (PLCC) as evaluation metrics.

Table \ref{tb:main_result} shows the comparison of the proposed method with the conventional NR-IQA methods such as BIQI \cite{Moorthy2020-biqi}, BLIINDS-II\cite{Saad2012-blinds2}, BRISQUE \cite{Mittal2012-wz}, and CLIP-IQA \cite{Wang2022exploring}. Some of results of the conventional methods are adopted from \cite{Wang2022exploring}.
From this result, we can see that the proposed method clearly outperforms the conventional NR-IQA methods in terms of both SROCC and PLCC. Note that CLIP-IQA is a special case of our proposed method. If we choose $N=1$ and $[\t_{1,1}, \t_{1,2}]=[$``Good photo.'', ``Bad photo.''$]$ in the proposed method, it is identical to CLIP-IQA.

\begin{table}[hbtp]
\caption{Comparison to other NR-IQA methods without task-specific training.The best results are shown in bold, and the second best results are underlined.}
    \label{tb:main_result}
    \centering
    \begin{tabular}{c|c|c}
        \hline
        Methods  & SROCC$\uparrow$  &  PLCC$\uparrow$  \\
        \hline 
        BIQI \cite{Moorthy2020-biqi}& 0.559  & 0.616 \\
        BLIINDS-II \cite{Saad2012-blinds2} & 0.585 & 0.598\\
        BRISQUE \cite{Mittal2012-wz} & \underline{0.705}  & 0.707 \\
        CLIP-IQA \cite{Wang2022exploring} & 0.695  & \underline{0.727} \\
        \bf{Ours} & \bf{0.719}  & \bf{0.752} \\
        \hline
    \end{tabular}
\end{table}

\begin{figure}[hbtp]
    \centering

  \begin{subfigure}[t]{3cm}
    \centering
    \includegraphics[width=\linewidth]{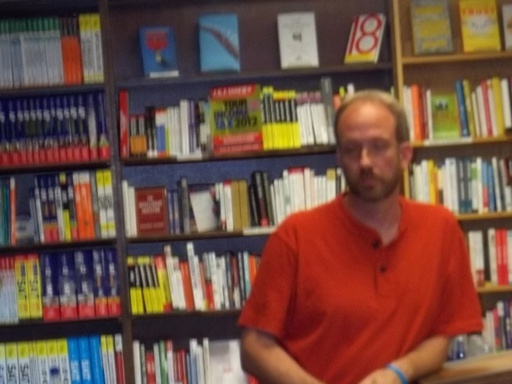}
    \subcaption{$d_1$=0.07, $d_3$=32.08, overall score $q$=13.32, MOS=19.09}
  \end{subfigure}
  %---
  \begin{subfigure}[t]{3cm}
    \centering
    \includegraphics[width=\linewidth]{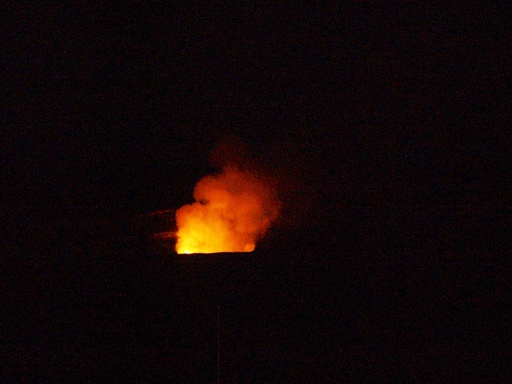}
    \caption{$d_1$=15.50, $d_3$=3.23, overall score $q$=14.04, MOS=20.13}
  \end{subfigure}

  \medskip

  \begin{subfigure}[t]{3cm}
    \centering
    \includegraphics[width=\linewidth]{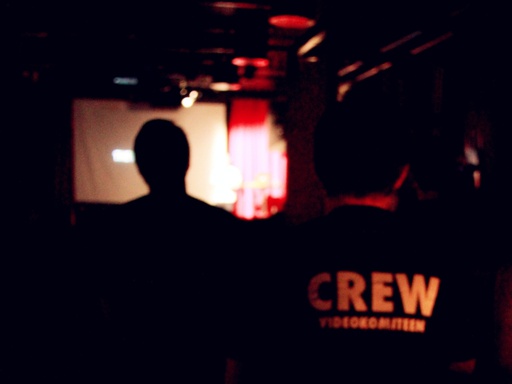}
    \caption{$d_1$=4.18, $d_3$=4.85, overall score $q$=16.35, MOS=18.69}
  \end{subfigure}
  %---
  \begin{subfigure}[t]{3cm}
    \centering
    \includegraphics[width=\linewidth]{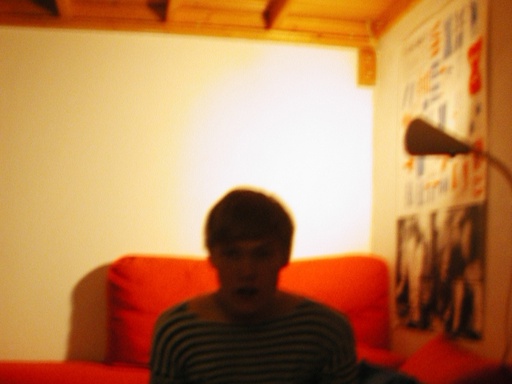}
    \caption{$d_1$=0.31, $d_3$=56.59, overall score $q$=23.32, MOS=5.50}
  \end{subfigure}
  %---
  \caption{Example of images and predicted scores. Note that $d_1$ and $d_3$ are the descriptive scores, which is from our proposed method, corresponds to sharpness and brightness.}
  \label{fig:description_capability}
\end{figure}

Fig. \ref{fig:description_capability} shows an example of the predicted scores corresponding to the images from the KonIQ-10k dataset.
Both of Fig. \ref{fig:description_capability}(a) and (b) show the images with low predicted scores and low MOS.  However, the cause of the image quality degradation is different. 
For example, Fig. \ref{fig:description_capability}(a) has a low score in sharpness, while Fig. \ref{fig:description_capability}(b) has a low score in brightness. 
Fig. \ref{fig:description_capability}(c) shows the image with low quality in both of sharpness and blightness. 
Fig. \ref{fig:description_capability}(d) is a failure case of the proposed method. The image has a low score in sharpness, but the MOS is high. This is because the proposed method estimates that this image has high MOS due to the brightness. However, the subject of this image is actually dark and the true MOS is low.

\section{Conclusion}
In this paper, we have proposed a zero-shot and interpretable IQA method using several descriptive features that may be related to the perceived quality of an image.
By developing multiple antonym-prompt pairs corresponding to carefully selected descriptive features and evaluating the similarity between them and the image using CLIP, it is possible to identify not only the perceived quality rating of the image, but also the cause on which the rating is based.
Experimental results show that the proposed method outperforms existing IQA methods in terms of estimation accuracy and is able to evaluate the causes of perceptual quality degradation.
The future work is to investigate the capability of the other prompt pairs.

\bibliographystyle{IEEEtran_miyata}
\bibliography{IEEEabrv,refs}

\end{document}